\documentclass[conference]{ieeeconf}
\usepackage{iftex}
\usepackage{amsmath,amssymb,amsfonts}
\usepackage{cite}
\usepackage{graphicx}
\usepackage{booktabs}
\usepackage{multirow}
\usepackage{xcolor}
\usepackage{url}
\usepackage{microtype}
\usepackage{placeins} 
\newcommand{\citep}[1]{\cite{#1}}

\graphicspath{{figures/}{.}}

\title{TEAR: Task-Evidence Authority Routing for Robotic Manipulation}
\title{TEAR: Task-Evidence Authority Routing for Robotic Manipulation}

\author{
\begingroup

\begin{tabular}{@{}ccc@{}}
Shicheng Zhang & Hui Liu & Haoliang Li \\
\multicolumn{3}{c}{Department of Electrical Engineering} \\
\multicolumn{3}{c}{City University of Hong Kong} \\
\texttt{\scriptsize szhang2578-c@my.cityu.edu.hk} &
\texttt{\scriptsize liuhui3-c@my.cityu.edu.hk} &
\texttt{\scriptsize haoliang.li@cityu.edu.hk}
\end{tabular}
\endgroup
}

\begin{document}
\maketitle
\begin{abstract}

Reliable robotic manipulation in field-deployed settings requires more than
executing a capable controller. As the execution conditions change, the system
must decide, online and on the robot side, which control source should hold
authority. Human teleoperation adapts readily to contact-rich and
semantically ambiguous situations but depends on a stable communication
link. Local autonomy is executed without continuous human intervention, but may
fail under semantic ambiguity, long-horizon drift, and unforeseen
contingencies. These two control sources exhibit complementary reliability
profiles.
We formulate this complementarity as a control-authority routing problem and
present \textbf{Task-Evidence Authority Routing for Robotic Manipulation
(TEAR)}, an edge-deployable framework that allocates authority between human
teleoperation and local autonomy. TEAR arbitrates on the robot side using
network-state evidence, which governs the usability of teleoperation, and
task-semantic evidence, which reflects whether autonomous execution remains
consistent with the current task. A lightweight temporal arbitrator observes
both forms of evidence together with robot state and current authority to
time authority transfers.
Across simulated and real-world trials under dynamic network conditions,
TEAR achieves task success rates close to pure teleoperation while limiting teleoperation to 19.8--27.4\% of control steps and improves success by 30--40\%  over
autonomy-only execution in our settings. These results demonstrate the
potential of complementary authority routing to sustain successful
manipulation with substantially reduced dependence on continuous human
control.

\end{abstract}

\section{Introduction}
\label{sec:intro}
Robotic manipulation~\citep{sheridan1992telerobotics,hokayem2006bilateral,niemeyer2004wave} is increasingly deployed beyond structured laboratory settings, in production-oriented and field-deployed environments where both communication quality and task conditions can vary over time. In such settings, reliability depends not only on the capability of the robot controller, but also on which source of control is in charge as execution conditions change. 
Human teleoperation remains one of the most dependable mechanisms for contact-rich and semantically ambiguous manipulation, because human experts can immediately adapt to unexpected contacts, partial observability, object displacement, and other contingencies. This reliability, however, is conditional on a stable communication loop. Classical teleoperation research has shown that latency and packet loss fundamentally affect transparency, stability, and user performance. 
More recent networked robotics studies further emphasise that latency, jitter, and packet loss remain practical bottlenecks for remote operation systems~\citep{kamtam2024latency,kizilkaya2024}. When the link deteriorates, the human may still know what should be done, but the robot no longer receives commands.

\begin{figure*}[t]
    \centering
    \includegraphics[width=\textwidth]{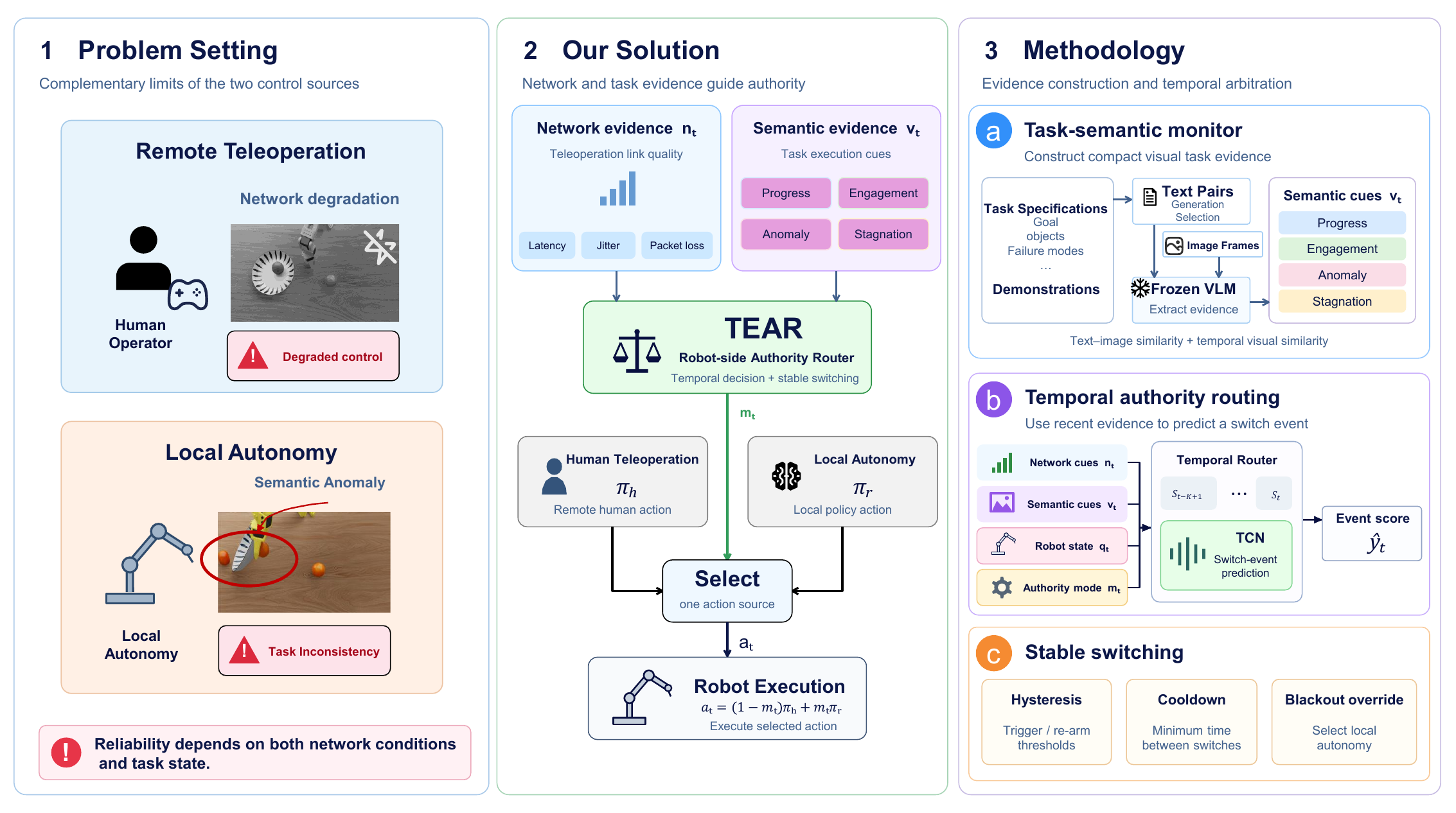}
    \caption{Overview of TEAR.
\textbf{(1) Problem setting.} Remote teleoperation and local autonomy
exhibit different failure conditions. Network degradation affects the
timeliness and continuity of human commands, while task inconsistency,
stagnation, or semantic anomalies can affect autonomous execution.
\textbf{(2) Our solution.} TEAR introduces a robot-side authority router
that jointly considers network and task evidence and assigns control to
either human teleoperation or the local VLA policy. The selected source
provides the action executed by the robot.
\textbf{(3) Methodology.} Task specifications and demonstrations guide offline
text-pair selection for a frozen VLM, which extracts progress,
engagement, anomaly, and stagnation cues from image observations.
These task cues are combined with network conditions, robot state, and
current authority to construct a temporal history for handoff-event
prediction. A causal temporal router predicts upcoming handoffs, and a
stateful gate with hysteresis and cooldown converts the event scores
into discrete authority transfers.}
    \label{fig:overview}
\end{figure*}

A direct response to this communication bottleneck is to reduce dependence on the remote operator through autonomous robot policies. Modern imitation learning and foundation model policies can map visual observations and task instructions to robot actions, and recent vision language action (VLA) models further transfer semantic knowledge from large scale vision language pretraining to robotic control~\citep{chi2023diffusion,zitkovich2023rt2,kim2024openvla,black2024pi0,nvidia2025gr00t}. Because such policies execute locally on the robot, they no longer depend on continuous human commands or on a stable command link, which directly addresses the failure mode described above. Autonomy alone, however, does not remove the need for human judgment. Learned policies may still stall, drift, or fail under semantic ambiguity, distribution shift, and unforeseen task contingencies. The two control sources therefore fail in different ways. Teleoperation remains adaptive but network dependent, and autonomous execution is locally available but semantically brittle.



Since neither control source alone can sustain reliable execution under
changing communication and task conditions, prior work has studied how
human and autonomous control can be coordinated. Classical shared autonomy
blends human and robot commands or infers the operator's goal to assist
task execution~\citep{dragan2013,javdani2018shared,herlant2016,jain2019}.
Mode-switching frameworks further automate transitions between autonomous
and teleoperated modes while accounting for communication
cost~\citep{kizilkaya2024}. Recent delay-aware shared autonomy adjusts the
contributions of human and autonomous control using teleoperation
transparency and autonomy confidence under network
delay~\citep{guleccyuz2025enhancing}. Policy uncertainty has also been used
to determine when human intervention is required during autonomous
execution~\citep{hoque2022thrifty}.

Our target setting further considers changes in both communication
quality and autonomous task execution during manipulation. The remote
command stream can be affected by latency, jitter, and packet loss, while
the local autonomous policy may encounter stalled progress, weak
engagement with the target, or undesired task states. Authority allocation
therefore requires evidence about the usability of remote control and the
current state of autonomous execution. The decision must also remain
available on the robot side as communication conditions change.
Motivated by this setting, we formulate control allocation as an
authority-routing problem between remote human teleoperation and an
independently executable local policy. The router observes both network
and task evidence over time and determines when authority should be
transferred between control courses.

We therefore present \textbf{Task-Evidence Authority Routing} (TEAR), an
edge-deployable framework for robot-side authority routing between human
teleoperation and local VLA control. First, TEAR combines network-state
evidence with task-semantic evidence. Network evidence characterizes the
usability of remote teleoperation, while task-semantic evidence captures
the execution state of local autonomy using a frozen vision-language
model and task-specific text probes prepared offline from the task
specification and demonstrations. Second, a causal temporal convolutional
network predicts upcoming handoffs from the recent history of network
conditions, task-semantic cues, robot proprioception, and current
authority. Third, a stateful gate with hysteresis and cooldown converts
predicted events into discrete authority transfers. Evidence estimation
and routing run locally without online queries from external models.

Across simulation and real-world experiments under time-varying network
conditions, we demonstrate that TEAR achieves task success rates close to
continuous teleoperation while using teleoperation for only
19.8--27.4\% of control steps, improving success by 30--40\%
points over autonomy-only execution in our evaluated settings. In summary, our contributions are:
\begin{itemize}
    \item We introduce TEAR, an edge-deployable framework that jointly uses
    network-state and task-semantic evidence to predict handoff events over
    time, and converts these predictions into stable authority transfers
    through a stateful gate with hysteresis and cooldown.

    \item We evaluate TEAR in simulated and real-world manipulation under
    time-varying network impairment, demonstrating its task success close to
    continuous teleoperation with substantially less human control, and
    analyze the visual and network evidence associated with expert handoffs.
\end{itemize}

\section{Related Work}
\label{sec:related_work}
\textbf{Teleoperation for Remote Manipulation and Robot Data Collection. }
Teleoperation remains a core paradigm for remote robotic manipulation, as
it transfers human perception, intent, and corrective judgment to robots
operating in hazardous, remote, or production-oriented environments.
Teleoperated robots have been widely used in disaster response, inspection,
industrial operation, and other field-deployed settings where full autonomy is
difficult to guarantee~\citep{murphy2014disaster,delmerico2019rescue,moniruzzaman2022teleoperation}.
Recent interfaces, including VR and low-cost leader--follower systems, further
make teleoperation an effective tool for collecting robot demonstrations and
training visuomotor policies~\citep{mandlekar2018roboturk,zhang2018vrteleop,zhao2023act}.
However, teleoperation is inherently constrained by the communication loop.
Latency, jitter, bandwidth variation, and packet loss affect both operator
perception and command execution, motivating recent studies on networked and
latency-aware teleoperation~\citep{kamtam2024latency,bray2024latency,kizilkaya2024}.
These communication constraints motivate using link conditions to guide control
allocation while retaining a locally executable manipulation policy.
TEAR builds on this observation by treating network state not merely as an
external disturbance, but as a decision signal for routing control authority.


\textbf{Autonomous Robot Policies. }
Autonomous robot policies aim to reduce dependence on continuous human control
by mapping observations and task specifications to executable actions.
Imitation learning learns such policies from expert demonstrations, with
methods such as DAgger addressing distribution shift through iterative data
aggregation~\citep{ross2011dagger}. Recent manipulation policies use
transformers, action chunking, and diffusion-based action generation to model
long-horizon or multimodal behaviours from demonstrations
~\citep{zhao2023act,chi2023diffusion}. Reinforcement learning provides another
route to autonomous control by optimizing interaction-driven objectives, as
shown in scalable grasping and continuous-control systems
~\citep{kalashnikov2018qtopt,haarnoja2018sac}. Model-based and world-model
approaches further learn predictive representations for planning and policy
learning~\citep{wu2023daydreamer}. More recently, vision-language-action
policies incorporate visual observations and language instructions into robot
control, enabling semantically conditioned manipulation across diverse tasks
~\citep{brohan2022rt1,zitkovich2023rt2,oneill2023openx,kim2024openvla,black2024pi0}.
These works show that autonomous policies are increasingly capable of local
perception and action. 
TEAR uses a locally deployed VLA as its autonomous controller. Runtime monitoring
provides complementary information about whether execution consistency checks with VLM-based
detection of task progression failures~\cite{agia2024sentinel}. In TEAR, visual task evidence forms part of 
the state used to decide when control should pass between the local policy and the human operator.


\textbf{Shared Autonomy and Mode Switching. }
Human-autonomy coordination has been studied through shared autonomy, assistive
teleoperation, and mode-switching systems. Classical shared-autonomy methods
infer user intent and blend human commands with autonomous assistance to
improve efficiency and reduce operator burden~\citep{dragan2013,javdani2018shared}.
Other work studies automatic mode selection for assistive teleoperation,
including time-optimal mode switching and latent action spaces for controlling
high-dimensional robots through lower-dimensional interfaces
~\citep{herlant2016,jeon2020latent,losey2021latent}. 
Recent systems extend
this direction with language and foundation models. LAMS uses large language
models to infer task context and switch assistive teleoperation modes, while
other LLM-assisted teleoperation systems use language models to interpret
operator intent or coordinate robot behaviours~\citep{tao2025lams,fei2025btla}.
Related work also explores coordination between teleoperation and VLA policies,
for example by assigning coarse motion to human teleoperation and fine-grained
manipulation to an autonomous policy~\citep{cui2025vla_shared_autonomy}, or by
using human interventions to refine robot policies after deployment
~\citep{xia2025preference}.
TEAR studies authority transfer under time-varying command impairment using
network observations and visual task cues as inputs to a robot-side
temporal predictor. Expert handoffs supervise upcoming-event prediction,
and a stateful gate regulates when a prediction changes authority. This
formulation makes the recent execution history and the current control
source explicit in the routing decision. Our evaluation examines this
design through task outcomes and offline handoff analysis.


\section{Method}
\label{sec:method}
TEAR considers remote robotic manipulation under network degradation and task-execution disturbances such as stalling, where human teleoperation and a locally deployed VLA policy provide complementary sources of control. Human teleoperation offers adaptive correction when autonomous execution stalls or encounters unexpected task conditions, while local autonomy provides an execution path when degraded communication disrupts the delivery of human commands. Within this setting, authority routing aims to improve manipulation resilience by assigning control according to the evolving conditions of both the communication link and task execution. Figure~\ref{fig:overview} presents our proposed framework, which integrates network and task evidence, predicts handoff events from their temporal history, and converts these predictions into discrete authority transfers. We first define the routing problem, then describe evidence construction, temporal prediction, and handoff execution.

\subsection{Problem Setup and Routing State}
\label{sec:setup}

Let $t$ index router updates and $m_t\in\{0,1\}$ denote current
authority. Teleoperation corresponds to $m_t=0$, and local VLA control
corresponds to $m_t=1$. The router selects $m_{t+1}$ for the next
execution interval. Within that interval, the selected authority $m$
determines the robot action as
\begin{equation}
 a=(1-m)a_h+ma_r,
\end{equation}
where $a_h$ is the available remote command and $a_r$ is the local
policy action. Each source has exclusive control during its interval.
The action-generation mechanisms of both controllers remain unchanged.

The routing state is
$\boldsymbol{s}_t=[\boldsymbol{n}_t^{\mathsf T},
\boldsymbol{q}_t^{\mathsf T},m_t,
\boldsymbol{v}_t^{\mathsf T}]^{\mathsf T}\in\mathbb R^{n_q+8}$,
where $\boldsymbol{n}_t\in\mathbb R^3$ contains network evidence,
$\boldsymbol{q}_t\in\mathbb R^{n_q}$ contains $n_q$ joint positions,
and $\boldsymbol{v}_t\in[0,1]^4$ contains visual task cues.
Continuous features are normalized using training-set statistics.
Authority remains binary.

\subsection{Complementary Evidence Construction}
\label{sec:evidence}

Network evidence
$\boldsymbol{n}_t=[\ell_t,j_t,d_t]^{\mathsf T}$
captures latency, jitter, and packet loss in the remote command stream.
Measurement details are given in Section~\ref{sec:protocol}.
Visual evidence
$\boldsymbol{v}_t=[v_t^{\mathrm{prog}},v_t^{\mathrm{stall}},
v_t^{\mathrm{eng}},v_t^{\mathrm{anom}}]^{\mathsf T}$
captures goal progress, visual stagnation, engagement between the target
and gripper, and specified undesired states. The cues are interpreted
jointly. High stagnation with high progress may indicate completion.
High stagnation with low progress may indicate a stalled attempt.

A task specification defines the objects, desired outcome, and relevant
failure states. Offline text generation proposes positive and negative
text pairs for progress, engagement, and anomaly. Task-specific labeled
demonstrations guide candidate selection. The selected pairs and
vision-language model (VLM) are fixed before deployment.
Text embeddings are precomputed and cached.

Let $\nu$ index visual updates. The unit-normalized image embedding
is $\boldsymbol{z}_\nu$, and the unit-normalized text embeddings are
$\boldsymbol{u}^{c,+}$ and $\boldsymbol{u}^{c,-}$ for cue
$c\in\{\mathrm{prog},\mathrm{eng},\mathrm{anom}\}$.
The positive text describes the presence of the cue.
For anomaly, it describes an undesired state.
The frozen VLM computes
\begin{equation}
 r_\nu^c=
 \frac{
 \exp(\lambda\boldsymbol{z}_\nu^{\mathsf T}\boldsymbol{u}^{c,+})
 }{
 \sum_{b\in\{+,-\}}
 \exp(\lambda\boldsymbol{z}_\nu^{\mathsf T}\boldsymbol{u}^{c,b})
 },
 \label{eq:vlm}
\end{equation}
where $\lambda>0$ is the encoder's logit scale. This scoring interface
supports CLIP~\cite{radford2021clip} and SigLIP~\cite{zhai2023siglip}.
The scores express relative agreement with the selected text pairs.
They do not directly estimate physical contact or failure probability.
Causal smoothing and persistence filtering suppress transient visual
changes before the scores enter the routing state.

Stagnation uses the same image embeddings. Let $\rho_\nu$ be the
fraction of consecutive-embedding comparisons within a recent visual
window whose cosine similarity exceeds $\eta$.
For a stagnation threshold $\rho_0\in[0,1)$, the stagnation cue is
\begin{equation}
 v_\nu^{\rm stall}=
 \max\!\left(0,\frac{\rho_\nu-\rho_0}{1-\rho_0}\right).
 \label{eq:stall}
\end{equation}
This cue remains zero until sufficient history is available.
Visual scoring, routing, and low-level control operate asynchronously.
At each router update, $\boldsymbol{v}_t$ contains the latest available
visual outputs. Evidence estimation and routing run locally without
online text generation or remote model queries.

\subsection{Learning Sparse Handoff Events}
\label{sec:events}

TEAR learns to predict upcoming handoffs from expert demonstrations.
For expert handoff annotations $e_t^{\rm D}\in\{0,1\}$, the training
target is
\begin{equation}
 y_t=\mathbb I\!\left[
 \sum_{h=1}^{H}e_{t+h}^{\rm D}>0
 \right],
 \label{eq:target}
\end{equation}
where $H$ is the future horizon in router samples. Near the end of an
episode, the sum includes only available samples from that episode.
The target identifies an upcoming handoff in either direction.
Future annotations are used only to construct training targets. This construction expands each annotated handoff into a short
pre-handoff supervision window of $H$ router updates. It allows the
predictor to learn evidence that develops before the annotated
authority transfer while preserving causal inference at deployment.

A causal temporal convolutional network (TCN)~\cite{bai2018tcn}
maps the preceding $K$ routing states to an event score
\begin{equation}
 \hat y_t=\sigma\!\left(
 f_\theta(\widetilde{\boldsymbol{s}}_{t-K+1:t})
 \right),
 \label{eq:event_score}
\end{equation}
where the tilde denotes feature normalization, $\theta$ denotes learned
parameters, and $\sigma$ is the logistic function. The TCN contains
three residual blocks with causal dilated convolutions, followed by
temporal pooling and a scalar output layer. Initial windows repeat the
first available state. Focal loss~\cite{lin2017focal} addresses the
imbalance between event-related and ordinary samples.

The same event predictor is used for both transfer directions.
Current authority $m_t$ identifies the active control source and
provides context for interpreting the observed evidence. An accepted
event under $m_t=0$ transfers authority from teleoperation to the local
VLA, while an accepted event under $m_t=1$ transfers authority to
teleoperation. The predictor therefore estimates handoff timing, with
the current authority determining the transfer direction.



\subsection{From Event Scores to Authority Transfers}
\label{sec:gate}

The event score is converted into an authority transfer by a stateful
gate. An armed gate accepts an event when
$\hat y_t \geq g_{\rm enter}$ and the cooldown is inactive.
After acceptance, the gate is disarmed and a cooldown of $L$ router
updates begins. The gate is re-armed once
$\hat y_t < g_{\rm exit}$, with
$g_{\rm exit} < g_{\rm enter}$. Re-arming may occur during cooldown,
although another transfer is allowed only after the cooldown expires.

For an accepted event $e_t\in\{0,1\}$, authority evolves as
\begin{equation}
m_{t+1}=
\begin{cases}
1-m_t, & e_t=1,\\
m_t, & e_t=0.
\end{cases}
\label{eq:switch}
\end{equation}

The two thresholds prevent repeated triggering from a sustained event
score, while the cooldown limits the frequency of successive transfers.
The active controller retains authority between accepted events.

\section{Experiments}
\label{sec:experiments}

We evaluate TEAR in simulation and real-world manipulation.
The experiments examine three aspects of the framework.
We first characterize teleoperation under network degradation.
We then evaluate task-level performance and teleoperation use.
Finally, we analyze the routing evidence and temporal handoff prediction.


\subsection{Experimental Setup}
\label{sec:protocol}

Simulation uses a SO-101 arm in Isaac Sim/IsaacLab~\cite{mittal2025isaaclab} to pick up
oranges from a table to a plate. The local controller is GR00T-N1.5~\cite{nvidia2025gr00t},
fine-tuned on 50 demonstrations. Real-world experiments use an SO-101
follower, a remote leader arm, and two RGB cameras to place yellow ducks
onto a plate. Ducks start at varying poses; successful execution places all
targets on the plate and stops the arm. The local controller is a separately
fine-tuned GR00T-N1.7-3B model~\cite{nvidia2026gr00t17}. The two tasks share
the pick-and-place structure, allowing us to examine the same
routing design across changes in objects, policies, and deployment conditions.

Remote commands are subject to software-imposed latency, jitter, and
packet loss. In the real-world controller, the latest command is used while
it remains fresh. Once it becomes stale, no new target is issued and the
position controller holds its last goal. Local VLA actions bypass the
impaired command path. Task-level comparisons use the time-varying impairment schedule
described below, with separate network realizations for each method.

Network evidence uses the imposed impairment parameters in simulation.
For real-world execution, latency and jitter are estimated from
receiver-side timing and expressed as round-trip time (RTT) by doubling
the one-way estimates. Packet loss is estimated from gaps in received
command sequences and represents missing command updates, including
application-level losses. For task-level evaluation, network impairment follows a predefined
time-varying schedule with three nominal levels. The corresponding
RTT settings are 50, 250, and 500~ms, jitter settings are 4, 50,
and 100~ms, and packet-loss settings are 0\%, 25\%, and 100\%.
Small fluctuations are introduced around the nominal latency and
jitter settings. The impairment level is updated every 8~s.
The router operates at 60~Hz in simulation and 30~Hz in the
real-world system. Visual task evidence is updated at 1.5~Hz using
CLIP ViT-B/32, with the latest available visual cues used at each
router update. The authority gate uses $g_{\rm enter}=0.40$,
$g_{\rm exit}=0.30$, and a cooldown of $L=20$ router updates.
The real-world controller treats a remote command as stale after
500~ms. All settings are fixed before task-level evaluation.

Routers are developed separately for the two settings. Simulation uses
40 expert-controlled episodes, split into 30 training and 10 validation
episodes, with 10 additional episodes reserved for offline event testing.
Real-world development uses 30 valid episodes. All router splits are by episode. A separate
ten-episode orange record supports the visual-cue analysis, and the original
20 real-world expert episodes support the handoff-evidence analysis.

Task-level evaluation compares Always-Teleop, Always-VLA, and TEAR using
ten valid rollouts per method and setting. Valid failures and timeouts are
included; invalid shutdown records are excluded. Real-world TEAR trials
are supervised: the logs contain 18 learned-router events and four manual
switches across two episodes, including one terminal switch.


We report task success, teleoperation use, and successful completion
time. For episode $i$, teleoperation use is measured as the fraction
of low-level control steps executed under teleoperation
\begin{equation}
 C_i=\frac{N_i^{\rm teleop}}{N_i^{\rm total}},\qquad
 \bar C=\frac{1}{N}\sum_{i=1}^{N}C_i,
 \label{eq:cost}
\end{equation}
where $N$ is the number of valid evaluation episodes. The metric is
averaged over successful and failed episodes. Completion time is
reported as mean and sample standard deviation over successful
episodes. Success proportions are accompanied by Wilson 95\%
confidence intervals.


\begin{figure}[!t]
 \centering
 \includegraphics[width=0.9\columnwidth]{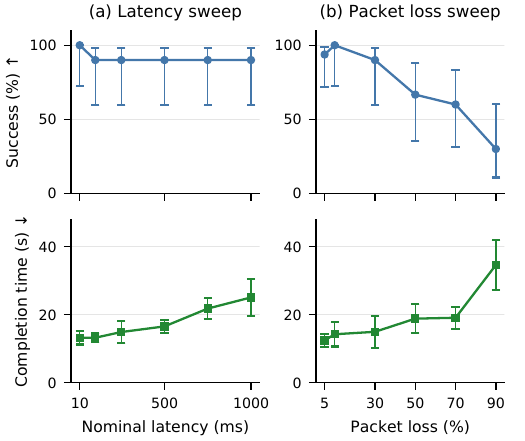}
 \caption{Teleoperation under fixed network impairment in simulation.
 (a) Nominal latency sweep. (b) Packet loss probability sweep. Success
 rates have Wilson 95\% confidence intervals; successful completion times
 show mean $\pm$ sample SD.}
 \label{fig:network}
\end{figure}

\begin{figure*}[!t]
 \centering
 \includegraphics[width=0.9\textwidth]{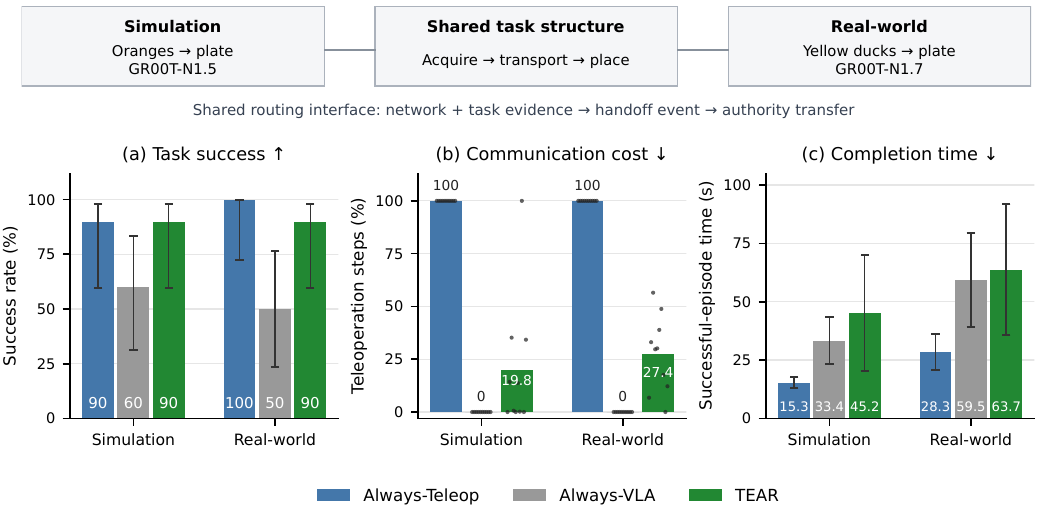}
 \caption{Task correspondence and outcomes across simulation and real-world
 deployment. The two settings share a manipulation structure and routing
 interface, with separately trained policies and routers. Each method has
 ten valid evaluation episodes per setting. (a) Success rates with Wilson
 95\% confidence intervals. (b) Mean episode-level communication cost.
 Dots show individual episodes, including failures. (c) Successful-episode
 completion time with sample-standard-deviation bars.}
 \label{fig:outcomes}
\end{figure*}

\subsection{Teleoperation under Network Degradation}
\label{sec:preliminary}

We first evaluate how network impairment affects teleoperation using
fixed-impairment orange trials. Figure~\ref{fig:network} reports
separate sweeps over nominal latency and packet loss. Increasing
latency mainly increases completion time among successful trials,
showing that delayed command delivery slows remote manipulation even
when the task can still be completed. At higher latency, the operator
must act through an increasingly delayed control loop, which reduces
the responsiveness of teleoperation.

Packet loss produces a different pattern. Moderate packet loss has a
limited effect on successful execution, while severe loss reduces the
observed success rate and also affects completion time. Missing command
updates reduce the continuity of the remote control stream and can
leave the robot operating on stale or incomplete commands. These
results show that the usability of teleoperation depends on both the
timeliness and continuity of remote commands. They provide the
communication-side motivation for including latency, jitter, and
packet-loss evidence in the TEAR routing state and for retaining a
locally executable VLA policy during degraded communication.

\subsection{Task Outcomes in Simulation and Real-World Deployment}
\label{sec:outcomes}



Figure~\ref{fig:outcomes} summarizes task-level performance in
simulation and real-world deployment. The two settings use different
objects, local VLA policies, and separately developed routers, while
sharing the same pick-and-place structure and authority-routing
interface. TEAR achieves 90\% task success in both settings.
Always-VLA achieves 60\% success in simulation and 50\% in the
real-world setting. Always-Teleop achieves 90\% and 100\%,
respectively. The results show that TEAR maintains task success close
to continuous teleoperation and improves over autonomous execution
alone in both settings.

TEAR also reduces the amount of teleoperation used during execution.
The mean teleoperation fraction is 19.8\% in simulation and 27.4\%
in the real-world trials. Relative to Always-Teleop, these values
correspond to reductions of 80.2\% and 72.6\%, respectively.
Most control steps under TEAR are therefore executed by the local VLA
policy, with teleoperation used during a smaller portion of each
episode. This pattern is observed in both the simulated and physical
robot settings.

Figure~\ref{fig:outcomes}(c) reports completion time for successful
episodes. In simulation, the mean completion times are 15.3~s for
Always-Teleop, 33.4~s for Always-VLA, and 45.2~s for TEAR.
In the real-world setting, the corresponding values are 28.3~s,
59.5~s, and 63.7~s. TEAR therefore requires longer completion time
than the two single-controller baselines in the evaluated successful
episodes. The task-level results show a consistent pattern across the
two settings, with lower teleoperation use than Always-Teleop and
higher task success than Always-VLA, together with increased
completion time.

The real-world TEAR trials are conducted under supervision as described
in Section~\ref{sec:protocol}. The recorded evaluation contains 18
learned-router events and four manual switches across two episodes,
including one terminal switch. The reported real-world task outcomes
therefore correspond to the supervised evaluation setting used in
these trials.




\subsection{Analysis of Routing Evidence}
\label{sec:semantic_results}

\begin{figure}[!tbp]
 \centering
 \includegraphics[width=0.8\columnwidth]{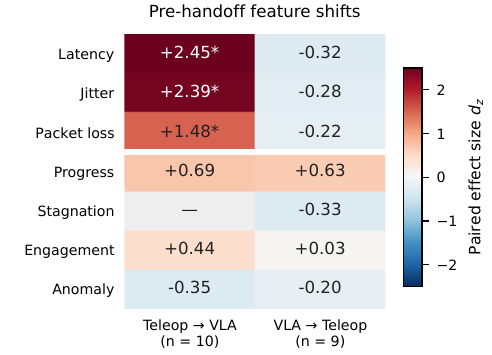}
 \caption{Real-world expert handoff evidence. Each cell shows the
 standardized paired difference $d_z$ (mean within-episode change divided
 by its sample SD), comparing pre-handoff and ordinary same-mode evidence.
 Episodes are the statistical units, with repeated same-direction events
 averaged within episode. Positive values indicate an increase before
 handoff. Asterisks mark FDR-adjusted $p<0.05$ across all 14 comparisons.
 The dash denotes identically zero paired changes.}
 \label{fig:real_handoff}
\end{figure}


We first examine how the visual task cues vary with task outcome.
The task-level rollouts use CLIP, while an offline SigLIP analysis
applies the same text-pair scoring interface with a second VLM encoder.
Table~\ref{tab:semantic} reports episode-level cue averages on the
calibration record, with each episode weighted equally within its
outcome group.

Successful episodes show higher mean progress and engagement and lower
mean stagnation and anomaly for both encoders. The absolute values vary
across CLIP and SigLIP, and several outputs approach saturation. The
direction of change is nevertheless consistent across the four cues in
this calibration record. This supports using the cues as complementary
inputs to the router rather than interpreting any individual score in
isolation. In particular, stagnation alone does not distinguish a
completed task from an unsuccessful stationary state. Its interpretation
depends on the concurrent progress and engagement evidence.

\begin{table}[!htbp]
 \centering\small
 \setlength{\tabcolsep}{3pt}
 \caption{Recorded visual cues on six successful (S) and four failed (F)
 orange episodes. Entries are means of episode means. Arrows indicate the
 intended direction for successful, advancing execution. This is a
 descriptive feature analysis on the calibration record.}
 \label{tab:semantic}
 \begin{tabular}{lrrrr}
 \toprule
 & \multicolumn{2}{c}{CLIP} & \multicolumn{2}{c}{SigLIP}\\
 \cmidrule(lr){2-3}\cmidrule(lr){4-5}
 Cue & S & F & S & F\\
 \midrule
 Progress $\uparrow$   & 0.532 & 0.485 & 0.500 & 0.448\\
 Stagnation $\downarrow$ & 0.000 & 0.134 & 0.000 & 0.119\\
 Engagement $\uparrow$ & 0.982 & 0.974 & 0.222 & 0.114\\
 Anomaly $\downarrow$  & 0.098 & 0.116 & 0.989 & 0.991\\
 \bottomrule
 \end{tabular}
\end{table}

We next examine how the full routing evidence changes around expert
handoffs. Figure~\ref{fig:real_handoff} compares the two seconds
preceding each handoff with ordinary execution under the same authority
mode. Reference frames exclude the three seconds surrounding a switch.
Repeated events in the same direction are averaged within each episode,
and statistical tests use false-discovery-rate correction across the
14 feature-by-direction comparisons.

The two transfer directions exhibit different evidence patterns.
Before Teleop-to-VLA handoffs, latency, jitter, and packet loss increase
relative to ordinary teleoperation periods. The corresponding effect
sizes are positive, and all three network features satisfy the adjusted
significance criterion. This is consistent with expert transfers away
from teleoperation occurring under degraded command delivery.

The evidence preceding VLA-to-Teleop handoffs is more heterogeneous.
The visual cues show smaller changes and none satisfies the same
significance criterion in the current demonstration corpus. Progress
increases before both transfer directions, while changes in stagnation,
engagement, and anomaly are weaker. These observations indicate that recovery from autonomous execution depends on the joint evolution of task evidence over time. The temporal router therefore integrates recent network, task, robot-state, and authority information to predict handoff events.

\subsection{Offline Handoff-Event Prediction}
\label{sec:event_results}
We evaluate temporal prediction on ten held-out simulation episodes
containing seven expert handoffs. Predicted events are matched
one-to-one with expert events within $\pm5$ router samples.
Event F1 summarizes precision and recall, while events per episode
measures prediction frequency. Recall is reported separately because
the handoff events are sparse and missed events remain relevant to the
routing behavior.

\begin{table}[!htbp]
 \centering\small
 \setlength{\tabcolsep}{4pt}
 \caption{Offline event prediction on ten test episodes. The TCN uses
 history $K=16$ and target horizon $H=10$. The rule is the highest-F1
 configuration in the evaluated sweep. Events/ep counts all predictions.}
 \label{tab:events}
 \begin{tabular}{lrrr}
 \toprule
 Predictor & Event F1 $\uparrow$ & Recall $\uparrow$ & Events/ep\\
 \midrule
 Rule sweep (best) & 0.171 & 0.429 & 2.8\\
 TCN & 0.308 & 0.286 & 0.6\\
 \bottomrule
 \end{tabular}
\end{table}
The TCN obtains an event F1 of 0.308 with 0.6 predicted events per
episode. The rule-based comparator obtains an event F1 of 0.171 with
2.8 predicted events per episode. The TCN therefore produces fewer
predicted events while achieving a higher F1 under the same held-out
episodes. Its recall is lower, with 0.286 compared with 0.429 for the
rule-based comparator, indicating that some expert handoffs are missed.

The difference in event frequency is relevant to the routing setting
because authority transfers are intended to be sparse. Frequent event
predictions can produce unnecessary opportunities for switching, while
a more selective predictor reduces the number of candidate transfers.
The lower recall of the TCN also shows that recovering infrequent expert
handoffs remains difficult with the current dataset. The result should
therefore be interpreted as evidence of a more selective event-prediction
pattern rather than complete recovery of expert switching behavior.

The offline analysis focuses on the temporal predictor independently
of the downstream authority-transfer logic. Predicted events are
extracted using threshold crossings with a cooldown, while deployed
routing applies the two-threshold gate described in
Section~\ref{sec:gate}. This separation provides a direct
characterization of the learned temporal handoff prediction used by
TEAR.

\FloatBarrier
\section{Limitations}
\label{sec:discussion}
\textbf{Task and deployment scope. }
The present evaluation focuses on two related pick-and-place tasks with
different objects, local VLA policies, and separately developed routers.
This setting allows the same authority-routing formulation to be examined
across simulation and physical robot execution. Broader task families,
additional operators, and more varied handoff situations can further
evaluate TEAR under a wider range of manipulation conditions.

\textbf{Network conditions. }
The current study uses controlled time-varying communication impairment
to provide repeatable changes in network quality. This setting does not
include all effects that may arise in operational networks, including
congestion, retransmission, correlated delay and loss, and coupled
degradation of command and video streams. Recorded and live network
traces can extend the evaluation to these conditions while retaining the
same network-evidence interface.

\textbf{Task evidence and routing configuration. }
The current task evidence uses only visual observations and focuses on
progress, stagnation, engagement, and anomaly. Fine contact and force
information is therefore outside the present evidence representation.
Additional sensory signals can be incorporated into the same routing
state to provide complementary task evidence. The current formulation
provides a compact interface for combining task-semantic evidence with
network conditions, robot state, and execution history for robot-side
authority routing.

\section{Conclusion}
\label{sec:conclusion}
We presented TEAR, a robot-side framework that coordinates human
teleoperation and local VLA control using complementary network and task
evidence. A frozen VLM provides visual task cues, and a causal temporal
router uses execution history to predict handoffs. Offline text
preparation, cached text embeddings, and local inference support edge
deployment without online generative-model queries for arbitration.
Across the evaluated simulation and real-world settings, TEAR achieves
task success rates close to those of pure teleoperation with substantially
less teleoperation use, at the cost of longer successful completion times.
These results highlight the practical potential of locally computed
authority routing for manipulation under changing communication and task
conditions.

\bibliographystyle{ieee_template/IEEEtran}
\bibliography{main}
\end{document}